\documentclass{article}
\usepackage[utf8]{inputenc}
\usepackage{graphicx}
\usepackage{amsmath}
\usepackage{amssymb}
\usepackage{textcomp}
\usepackage{wrapfig}
\usepackage{color}
\usepackage{mathtools}

\usepackage{pbox}

\usepackage[square,sort,comma,numbers]{natbib}

\usepackage[preprint]{neurips_2020}
\usepackage[utf8]{inputenc} % allow utf-8 input
\usepackage[T1]{fontenc}    % use 8-bit T1 fonts
\usepackage{hyperref}       % hyperlinks
\usepackage{url}            % simple URL typesetting
\usepackage{booktabs}       % professional-quality tables
\usepackage{amsfonts}       % blackboard math symbols
\usepackage{nicefrac}       % compact symbols for 1/2, etc.
\usepackage{microtype}      % microtypography

\usepackage[square,sort,comma,numbers]{natbib}

\title{Relevance of Rotationally Equivariant Convolutions for Predicting Molecular Properties}

\author{%
  Benjamin Kurt Miller \\
  Freie Universit\"at Berlin\\
  Berlin, Germany\\
  Lawrence Berkeley National Laboratory \\
  Berkeley, California, USA \\
  \texttt{b.k.miller@uva.nl} \\
   \And
   Mario Geiger \\
   EPFL \\
   Lausanne, Switzerland \\
   Lawrence Berkeley National Laboratory \\
   Berkeley, California, USA \\
   \texttt{mario.geiger@epfl.ch} \\
   \AND
   Tess E. Smidt \\
   Lawrence Berkeley National Laboratory \\
   Berkeley, California, USA \\
   \texttt{tsmidt@lbl.gov} \\
   \And
   Frank No\'e \\
   Freie Universit\"at Berlin \\
   Berlin, Germany \\
   Rice University \\
   Houston, Texas, USA \\
   \texttt{frank.noe@fu-berlin.de} \\
}

\date{April 2020}

\begin{document}

\maketitle

\begin{abstract}
    Equivariant neural networks (ENNs) are graph neural networks embedded in $\mathbb{R}^3$ and are well suited for predicting molecular properties. The ENN library \texttt{e3nn} has customizable convolutions, which can be designed to depend only on distances between points, or also on angular features, making them rotationally invariant, or equivariant, respectively. This paper studies the practical value of including angular dependencies for molecular property prediction directly via an ablation study with \texttt{e3nn} and the QM9 data set. We find that, for fixed network depth and parameter count, adding angular features decreased test error by an average of 23\%. Meanwhile, increasing network depth decreased test error by only 4\% on average, implying that rotationally equivariant layers are comparatively parameter efficient. We present an explanation of the accuracy improvement on the dipole moment, the target which benefited most from the introduction of angular features.
    % Similar results were found when the network depth was increased; however,
    % for increasing network depth. For most, but not all, molecular properties, distance-only \texttt{e3nn}s (L0Nets) can compensate by increasing convolutional layer depth. We argue that the parameter efficiency of our \texttt{e3nn} (L1Net) architecture comes from the rotation properties of its angular-features, in particular on the electronic dipole moment. 
    % Our angular-feature \texttt{e3nn} (L1Net) architecture beats previous state-of-the-art results on the global electronic properties dipole moment, isotropic polarizability, and electronic spatial extent.
\end{abstract}

\section{Introduction}

The discovery of novel molecules has been accelerated by advances in computational quantum chemistry and machine learning assisted exploration of chemical space \cite{halicin, robin_molecule_discovery, molecule_net}. The successes have been characterized by designing bespoke neural networks which have relevant properties ``baked-in,'' such as parameter sharing across calculations on individual atoms, continuous convolutions, invariance to atomic indexing, and invariance to rotation and translation \cite{schnet, ani1}. 
Meanwhile, there has also been development on neural networks which are equivariant to group action \cite{se3cnn}, some with molecules in mind \cite{tensorfieldnetworks, cormorant}.
% , i.e. $\forall g \in G, \forall x \in X, f(g \cdot x) = g \cdot f(x)$.
Equivariant neural networks can be seen as a super-set of invariant ones because a group necessarily contains the identity element.
% , i.e. an invariant function is an equivariant function which collapses the orbit of each element of its domain to the same value in the codomain, i.e. $\forall g \in G, \forall x \in X, f(g \cdot x) = f(x)$. 
The question considered in this paper can loosely be stated as: When doing regression on scalar molecular properties, what is missing when one employs only invariant layers in a neural network as opposed to including equivariant ones?

We explore this question using the QM9 benchmark \cite{gdb17, qm9} by predicting quantum chemical properties of small molecules. While the molecules can rotate and translate, affecting the molecule's position vectors, the QM9 properties are all scalar and invariant to translation or rotation. Here we compare equivariant neural networks (ENNs) that predict rototranslationally invariant molecular properties but differ by whether their internal features are rotationally invariant (convolutions depend on distances) or equivariant (convolutions depend on distances and angles). We also investigate whether increasing depth in networks with rotationally invariant layers is comparatively effective at reducing test error. The networks are implemented in the PyTorch \cite{pytorch} library e3nn 
\cite{e3nn}
%\cite{e3nn_paper, e3nn}
using the $SE(3)$ equivariant point modules. QM9 data handling and training routines were borrowed from SchNetPack \cite{schnetpack}.

Given atomic positions $\mathbf{r} \in \mathbb{R}^{3 \times N}$ and atomic features $F^{h}$, layer $h$ of an \texttt{e3nn} produces atomic features $F^{h+1}$ by $\mathcal{L}^{h}(\mathbf{r}, F^{h}) = F^{h+1}$. $F^{h}$ is a collection of $u_{0}^{h}$ scalars $F^{h}_{\ell=0}$ and $u_{1}^{h}$ vectors $F^{h}_{\ell=1}$ flattened into a column by $F^{h} = \text{vec}(F^{h}_{\ell=0} \oplus F^{h}_{\ell=1})$. The total multiplicity of features at layer $h$ is $u^{h} = u_{0}^{h} + u_{1}^{h}$. The rotation matrix $\mathbf{R}$ acts on $F^{h}$ in block matrix notation by 

\begin{equation}
    \mathbf{R} F^{h} = 
    \begin{pmatrix}
        \mathbf{R}_{\ell=0} & \mathbf{0} \\ 
        \mathbf{0} & \mathbf{R}_{\ell=1}
    \end{pmatrix}
    \begin{pmatrix}
        F^{h}_{\ell=0} \\ 
        F^{h}_{\ell=1}
    \end{pmatrix}
    =
    \begin{pmatrix}
        \mathbf{1} & \mathbf{0} \\ 
        \mathbf{0} & \mathbf{R}_{\ell=1}
    \end{pmatrix}
    \begin{pmatrix}
        F^{h}_{\ell=0} \\ 
        F^{h}_{\ell=1}
    \end{pmatrix}.
\end{equation}

In this paper, we consider the case of rotationally invariant layers, which produce features that \textit{do not rotate}, i.e. $u_{1} = 0$, and compare their performance to rotationally equivariant layers, which produce \textit{rotating} features, i.e. $u_{1} \neq 0$. In order to predict a rotationally invariant target value, the output features, $F^{h_{max}} = F^{h_{max}}_{\ell=0}$, do not rotate. The difference between networks lies in the equivariance or invariance of their internal layers. We call networks containing only features that do not rotate and rotationally invariant layers L0Nets, while networks containing rotating features and equivariant layers are called L1Nets. A more general framing in terms of spherical harmonics can be found in the \texttt{e3nn} library 
\cite{e3nn}
%\cite{e3nn_paper, e3nn}
. In said framing, features are seen as spherical harmonics of degree $\ell$.

\subsection{Related Work} 
Molecular properties, which depend only on the atomic distance graph, are commonly predicted by kernel methods or Gaussian process regression \cite{mol_kernel_0, mol_kernel_1, mol_kernel_2, mol_kernel_3} 
% \textcolor{red}{cite 3-5 relevant papers. Include seminal papers by Rupp/Mueller/Tkatchenko/von Lilienfeld, Czanyi and Ceriotti} 
or graph neural networks \cite{graphs_0, graph_1},
% \textcolor{red}{cite 3-5 relevant papers. Include: https://arxiv.org/abs/1806.01261    https://pubs.rsc.org/en/content/articlelanding/2019/sc/c8sc04175j#!divAbstract}
where ENNs are usually employed for predicting physical properties, which depend on atomic displacement vectors \cite{husic2020coarse, geometric_new_paper}.
% What does frank mean with this manipulation of the sentence? Not it doesn't make sense to me. It's repetitive.
% \textcolor{red}{cite a few examples, including https://arxiv.org/abs/2007.11412}
While kernel approaches are more data-efficient, graph neural networks scale to larger amounts of data. 
Inspiration for our study came from literature on invariant and equivariant ENNs for molecular property prediction. % this sentence could go
SchNet \cite{schnet, schnetpack} introduced atom-wise features with continuous convolution. Tensor Field Networks \cite{tensorfieldnetworks} and Cormorant \cite{cormorant} generalized the approach to angular-feature based rotation equivariant networks. In parallel, although aimed at voxelized data, se3cnn \cite{se3cnn} developed the gated nonlinearity. The library under consideration, \texttt{e3nn}, represents a superset of Tensor Field Networks, SchNet, and se3cnn. Support for Cormorant's so-called two-body interaction has also been included in \texttt{e3nn} but is not considered in this experiment.

Although DimeNet \cite{dimenet} is the leading architecture on QM9 regression, it is not considered in our analysis. Their use of directional message passing, while effective, is not trivially compatible with SchNet, Cormorant, or \texttt{e3nn}. Their edge featurization using spherical Fourier-Bessel functions could be incorporated rather simply, but investigation is left for future work.

While QM9 remains the gold standard for most machine learning studies, a new data set called QM7-X \cite{qm7x} contains a wealth of tensor properties suitable for prediction with networks like \texttt{e3nn} or Cormorant. SchNet and DimeNet cannot predict tensor quantities in their current incarnations.

% Table \ref{table:relatives} compares the features of these architectures.

% \begin{table}[htb]
%     \centering
%     	\resizebox{\textwidth}{!}{
%         \begin{tabular}{|c || c | c | c | c | c |} 
%             \hline
%              & Point Clouds & SO(3) Features & Gated Nonlinearity & Two-body Interaction\\
%             \hline\hline
%             e3nn (Tensor Field Networks 2.0) & \checkmark & \checkmark & \checkmark &  \\ 
%             \hline
%             SchNet & \checkmark & & & \\
%             \hline
%             Cormorant & \checkmark  & \checkmark & & \checkmark \\
%             \hline
%             se3cnn &  & \checkmark & \checkmark & \\
%             \hline
%         \end{tabular}
%     }
%     \caption{Features of ENN architectures. SchNet operates on point clouds, has layers which are invariant to rotation, and does not employ angular information internally. With the right hyperparameters, LONet can be made identical to SchNet. Cormorant uses spherical harmonic features, like e3nn, but has no nonlinearity; however, it includes a so-called two-body interaction. se3cnn introduced the gated nonlinearity, but operates on voxelized data. Tensor Field Networks used a norm nonlinearity.}
%     \label{table:relatives}
% \end{table}

\section{Methods}

\begin{figure}[htb]
    \centering
    \includegraphics[width=0.9\textwidth]{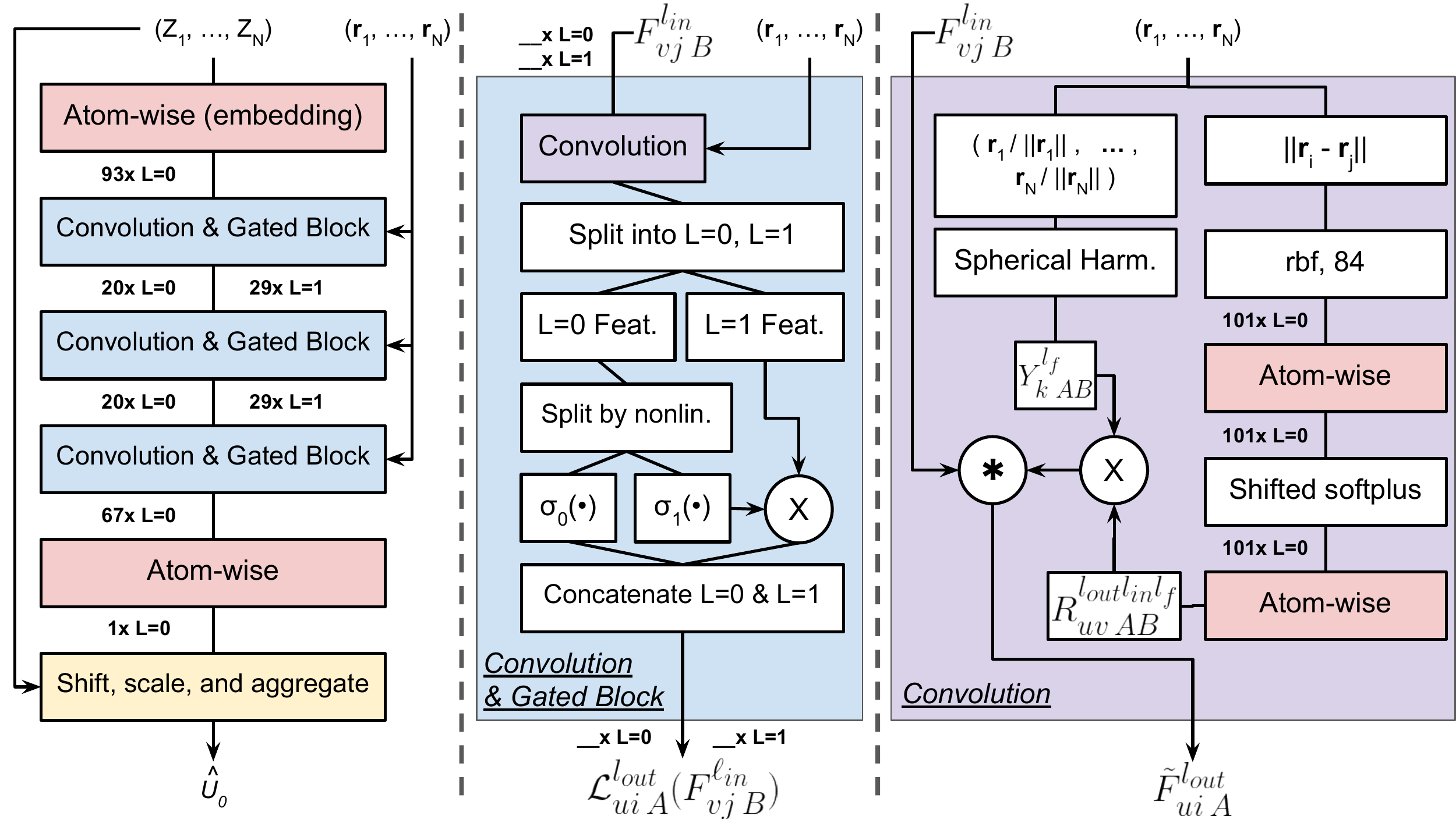}
    \caption{Illustration of L1Net with the architecture on the left, the convolution \& gated block in the center, and the convolution on the right. The scalar activation function $\sigma_{0}(\cdot) = \text{Softplus}(\cdot)$, while the gated activation function $\sigma_{1}(\cdot) = \text{Sigmoid}(x)$. The notation ``$T$ x L = $D$'' implies that this connection contains a multiplicity $T$ of features with degree $D$. The target output $\hat{U}_{0}$ is shown as an example.}
    \label{fig:e3nn_l0_architecture}
\end{figure}

We employ both an L0Net and an L1Net to do regression on scalar target values from the QM9 data set given molecular input data. A molecule is an unordered set of $N \in \mathbb{N}$ atoms, each with position $\mathbf{r}_{a} \in \mathbb{R}^{3}$ and element $Z_{a}$ which is represented as a one-hot scalar array. We parameterize a neural network such that $\{(\mathbf{r}_{1}, Z_{1}), ..., (\mathbf{r}_{N}, Z_{N}) \} \mapsto \mathcal{F}(\mathbf{r}_{1}, ..., \mathbf{r}_{N}, Z_{1}, ..., Z_{N})$ where we restrict the image to be invariant to rotations and translations, as well as permutations in atomic indexing. Every layer uses parameter sharing across atoms and the final step accumulates the value of every atom with a symmetric function. A schematic of the entire architecture can be seen in Figure \ref{fig:e3nn_l0_architecture}.

\subsubsubsection{\textbf{Atom-wise}} \hspace{1 mm}
A dense layer applied to every atom with parameters shared across atoms. Given weights $W_{u' u}$, bias $b_{u}$, non-rotating, scalar features $F$ on atom $A$ at layer $h$ with multiplicity $u'$ we write $F^{h+1}_{u\, A} = \sum_{u'} F^{h}_{u'\, A} W_{u' u} + b_{u'}$. This layer is also used as a learned embedding of the atomic element.

\subsubsubsection{\textbf{Radial Basis Function (rbf)}} \hspace{1 mm}
The radial basis $\phi: \mathbb{R} \to \mathbb{R}^{\mathcal{B}}$, expands $d = \lVert \mathbf{r}_{b} - \mathbf{r}_{a} \rVert$ by
\begin{equation}
	\label{eqn:cosine_basis}
    \phi(d)
	= 
	\begin{cases}
		cos^2 (\frac{\pi}{2} \frac{d - \mu_{\mathcal{B}}}{\mu_{\mathcal{B} + 1} - \mu_{\mathcal{B}}}) & -1 \leq \frac{d - \mu_{\mathcal{B}}}{\mu_{\mathcal{B} + 1} - \mu_{\mathcal{B}}} \leq 1 \\
		0 & \text{otherwise},
	\end{cases}
\end{equation}
where $0 \, \text{\AA} \leq \mu_{\mathcal{B}} \leq 11.1 \, \text{\AA}$ is a sequence of $\mathcal{B} = 84$ equally spaced ``radial basis centers.''

\subsubsubsection{\textbf{Convolution}} \hspace{1 mm}
The convolutional filter $f$ consists of a learned scalar array radial function multiplied by a multiplicity of spherical harmonics of degree $\ell_{f}$. The filter is assigned an atomic index based on the atom on which it was evaluated. The filter and the atomic features interact which necessitates a double atomic indexing of $A$ and $B$. The degree indices $\ell_{out}, \ell_{in}, \ell_{f}$ correspond to order indices $i, j, k$ respectively. $u, v$ both represent multiplicity. Using the input features $F$, Clebsch-Gordan coefficients $C$, filter spherical harmonics 
$Y(\frac{\mathbf{r}_{B} - \mathbf{r}_{A}}{\Vert \mathbf{r}_{B} - \mathbf{r}_{A} \Vert})$,
learned scalar radial coefficients $R(\phi(\lVert \mathbf{r}_{B} - \mathbf{r}_{A} \rVert))$, and normalization coefficients $n$, the convolutional output $\Tilde{F}$ is defined, with the layer $h$ index omitted,

\begin{equation}
	\label{eqn:kc_forward}
	\Tilde{F}^{\, \ell_{out}}_{ui\, A} 
	= \sum_{B\; \ell_{f}\; \ell_{in}\; vj\; k\;} C^{\ell_{out}\ell_{in}\ell_f}_{ijk} \; Y^{\ell_f}_{k\,AB} \; R^{\ell_{out}\ell_{in}\ell_{f}}_{uv\, AB} \; n^{\ell_{out}\ell_{in}}_{AB} \; F^{\ell_{in}}_{vj\, B}.
\end{equation}

The customization between the invariant, scalar-only, distance-based L0Net and the equivariant, scalar-and-vector, distance-and-angle-based L1Net is determined by the degrees $\ell_{in}, \ell_{out}$. L0Nets only use $\ell_{in}, \ell_{out} = 0$ while L1Nets allow for $\ell_{in}, \ell_{out} \in \{0, 1\}$.
The normalization is defined such that input features, with component-wise unity second moments, and component-wise normally distributed radial components produce features with component-wise unity second moments.

\subsubsubsection{\textbf{Gated Block}}  \hspace{1 mm}
This layer is used to provide a nonlinearity to the output of the convolution. Scalars are handled normally, i.e. $\mathcal{L}(F^{\ell=0}) = \text{Softplus}(\Tilde{F}^{\ell=0})$, while vector, $\ell=1$, features are multiplied by a scalar passed through an activation function. Specifically, $\mathcal{L}(F^{\ell=1}_{u}) = \text{Sigmoid}(\Tilde{F}^{\ell=0}_{u + \mathcal{O}}) \Tilde{F}^{\ell=1}_{u}$. This introduces nonlinearity while maintaining equivariance. The previous layer produces extra learned scalar features, of multiplicity $u_{1}$ with index offset $\mathcal{O}$, in order to utilize this nonlinearity.

\subsubsubsection{\textbf{Final Atom-wise and Shift, Scale, Aggregate}}  \hspace{1 mm} The last Convolution \& Gated Block is restricted to output scalar, non-rotating features, facilitating an atom-wise layer on those features while retaining overall rotation invariance. The final atomic features are summed to produce a single scalar output, $P = \sum_{a=0}^{N} F^{h_{max}}_{a}$.
In order to keep $P$ near mean zero and variance one, it is shifted and scaled using statistics calculated from the training set and atomic references from the QM9 data set to finally output the target prediction, $\hat{target}$. We employ the MSE loss between $\hat{target}$ and $target$.

\subsection{Experiment}
Using QM9 and following the training procedure from SchNetPack, we selected random training, validation and test sets with 109,000, 1,000 and 23,885 molecules, respectively.
Each network architecture was trained on each of the 12 QM9
properties. This procedure was repeated three times with an L1Net, an L0Net, and an L0Net Deep, where L0Net Deep has an additional Convolution \& Gated Block. The specifics of the three network architectures were determined by hyperparameter search, as described in the supplementary material. The L0Net is the same as the L1Net, except that the $F^{\ell=1}_{u=1,...,29}$ features are dropped and the multiplicity of $F^{\ell = 0}$ features is increased by $3 \times 29 = 87$.

The adam optimizer \cite{adam} was employed with standard parameters and an initial learning rate of $6.53 \times 10^{-3}$. The learning rate was exponentially decayed by factor 0.5 on a loss plateau of 5 epochs to a minimum of $10^{-7}$. Maximum training epochs was set at 200 with early stopping patience of 50.

\begin{table}[htb]
	\centering
	\resizebox{\textwidth}{!}{
        \begin{tabular}{l|rrr|rrr||rrrr}
        \toprule
                                           Target &  SchNet &  \pbox{2cm}{SchNet \\ Pack} &  Cormorant &   L1Net &   L0Net &  \pbox{2cm}{L0Net \\ Deep} &  \%$\mathcal{E}_{\text{L1}, \text{L0}}$ &  \%$\mathcal{E}_{\text{Deep}, \text{L0}}$ &  $\Delta \mathcal{E}_{\text{L1}, \text{Deep}}$ &  \%$\mathcal{E}_{\text{L1}, \text{Deep}}$ \\
        \midrule
                                        $\mu$ (D) &   0.033 &                       \textbf{0.021} &      0.038 &   \textbf{0.043} &   0.086 &                      0.091 &                                  -0.501 &                                     0.055 &                                         -0.048 &                                    -0.556 \\
                             $\alpha$ ($a_0^{3}$) &   0.235 &                       0.124 &      \textbf{0.085} &   \textbf{0.088} &   0.115 &                      0.115 &                                  -0.235 &                                     0.000 &                                         -0.027 &                                    -0.235 \\
                          $\epsilon_{HOMO}$ (meV) &  41.000 &                      47.000 &     \textbf{34.000} &  46.015 &  47.069 &                     \textbf{45.294} &                                  -0.022 &                                    -0.038 &                                          0.721 &                                     0.015 \\
                          $\epsilon_{LUMO}$ (meV) &  \textbf{34.000} &                      39.000 &     38.000 &  \textbf{34.646} &  39.947 &                     37.217 &                                  -0.133 &                                    -0.068 &                                         -2.571 &                                    -0.064 \\
                           $\epsilon_{gap}$ (meV) &  63.000 &                      74.000 &     \textbf{61.000} &  \textbf{67.543} &  70.344 &                     67.873 &                                  -0.040 &                                    -0.035 &                                         -0.330 &                                    -0.005 \\
              $\langle R^{2} \rangle$ ($a_0^{2}$) &   \textbf{0.073} &                       0.158 &      0.961 &   \textbf{0.354} &   0.579 &                      0.382 &                                  -0.389 &                                    -0.340 &                                         -0.028 &                                    -0.048 \\
                                       zpve (meV) &   1.700 &                       \textbf{1.616} &      2.027 &   \textbf{1.561} &   1.804 &                      1.800 &                                  -0.135 &                                    -0.002 &                                         -0.239 &                                    -0.132 \\
                                      $U_0$ (meV) &  14.000 &                      \textbf{12.000} &     22.000 &  \textbf{13.464} &  19.943 &                     18.487 &                                  -0.325 &                                    -0.073 &                                         -5.023 &                                    -0.252 \\
                                          U (meV) &  19.000 &                      \textbf{12.000} &     21.000 &  \textbf{13.834} &  19.889 &                     19.533 &                                  -0.304 &                                    -0.018 &                                         -5.699 &                                    -0.287 \\
                                          H (meV) &  14.000 &                      \textbf{12.000} &     21.000 &  \textbf{14.358} &  21.001 &                     20.744 &                                  -0.316 &                                    -0.012 &                                         -6.386 &                                    -0.304 \\
                                          G (meV) &  14.000 &                      \textbf{13.000} &     20.000 &  \textbf{13.989} &  20.057 &                     18.744 &                                  -0.303 &                                    -0.065 &                                         -4.755 &                                    -0.237 \\
         $C_v$ ($\frac{\text{cal}}{\text{molK}}$) &   0.033 &                       0.034 &      \textbf{0.026} &   \textbf{0.031} &   0.035 &                      0.037 &                                  -0.114 &                                     0.057 &                                         -0.006 &                                    -0.171 \\
        \bottomrule
        \end{tabular}
    }
    \caption{%
        This table quantifies the mean absolute error of relevant models on the QM9 regression targets over unseen test data. The L1 and L0Nets are compared to their closest relatives, SchNet and Cormorant as well as an L0Net with an extra Convolution \& Gated Block layer called L0Net Deep. $\Delta \mathcal{E}_{X, Y}$ implies $X - Y$ where $X, Y$ are mean absolute errors of models. \%$\mathcal{E}_{X, Y}$ is the same calculation divided by the L0Net mean absolute error on the same target. The size of train/validation/test sets differed across SchNet, SchNetPack, and Cormorant. The L$\cdot$Nets were trained like the published version of SchNetPack \cite{schnetpack} in this regard. Bold face indicates best performance within the sub-table.
    }%
    \label{table:performance}
    \vspace{-8mm}
\end{table}

\begin{wrapfigure}[12]{R}{0.3\textwidth}
  \vspace{-.5cm}
  \begin{center}
    \includegraphics[width=0.3\textwidth]{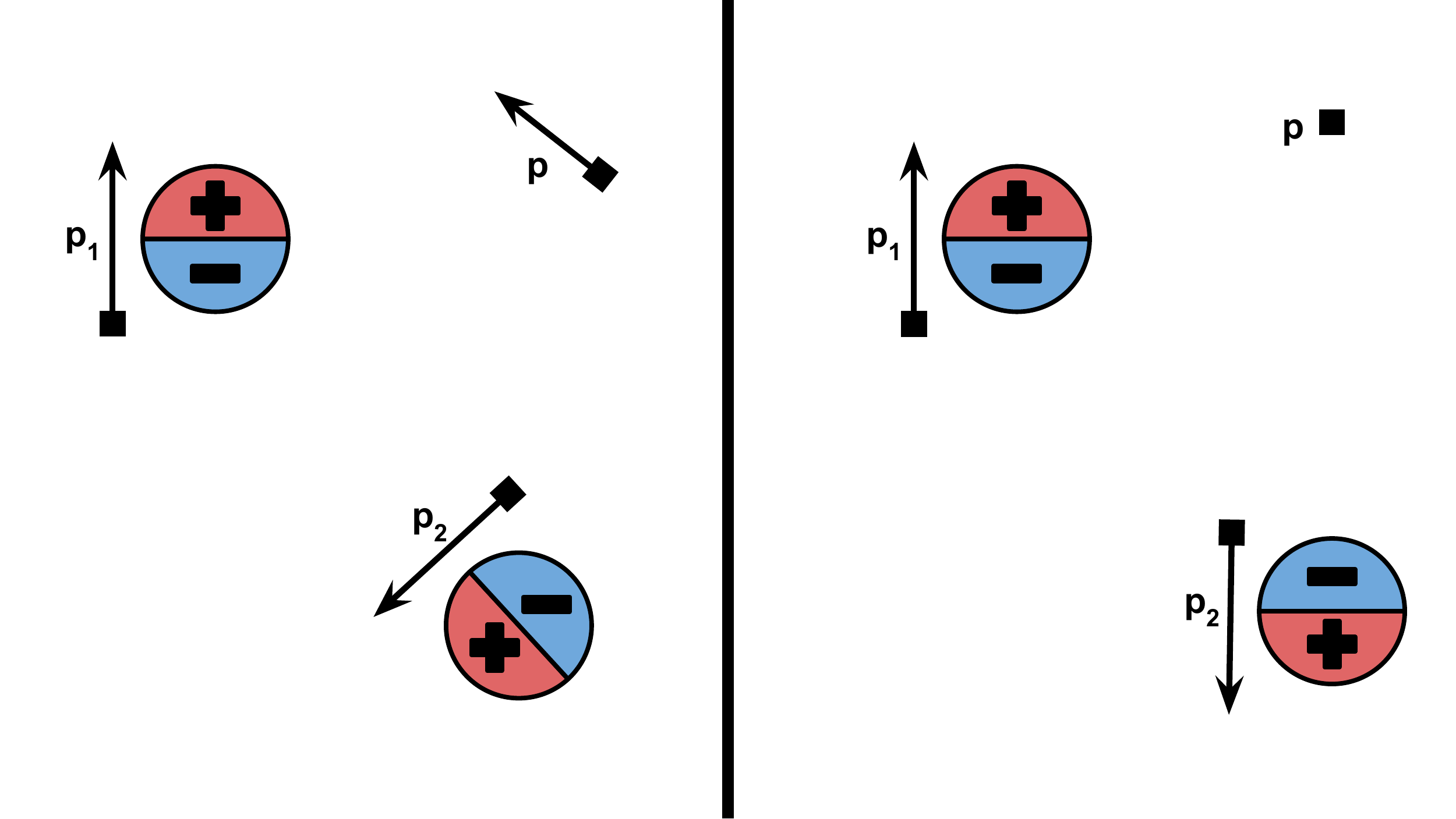}
  \end{center}
  \caption{The magnitude of the  total dipole moment depends on the orientation of the constituents, which L0Net convolutions do not consider.}
%   \vspace{15mm}
  \label{fig:dipole_addition}
\end{wrapfigure}

We quantify the performance across L0Net, L1Net, and L0Net Deep in Table \ref{table:performance}. If we average the \%$\mathcal{E}_{L1, L0}$ column across targets, we find that introducing rotating features improved performance by 23\% on the mean absolute error. In other words, the ablation of rotating features, by changing L1Net to L0Net without altering the architecture otherwise, significantly reduced the parameter efficiency by decreasing the accuracy and keeping the number of parameters constant.

\newpage

We compare these results to an alternative modification of the architecture, namely, introducing another, rotationally invariant, Convolution \& Gated Block to our L0Net. Averaging \%$\mathcal{E}_{Deep, L0}$ across targets reveals that L0Net Deep reduces the error by an average of 4\%--significantly less effective while introducing another layer of parameters. Notably, L1Net improved on every target when compared to L0Net, while L0Net Deep worsened predictions on  $C_{v}$ and $\mu$.

% Rather than proof, think of it as the "use-cases" which L1Net offers compared to L0Net. 
% I think the below intuition doesn't really apply to molecules anymore. Rather I want to emphasize that you get a much richer expressivity by adding l1 features instead of more l0 layers. (parameter efficiency)

In simple cases, predicting the magnitude of a rotating vector quantity, like $\mu$, requires functional dependence on the angles between constituents in order to make unbiased predictions. Consider the case of an estimator $\mathcal{F}$ which predicts the magnitude squared total dipole moment of two constituent dipoles $p^{2} = \lVert \mathbf{p}_1 + \mathbf{p}_2 \rVert^{2} = p_1^2 + 2 p_1 p_2 \cos{\theta_{12}} + p_2^2$. $\mathcal{F}$ is restricted from functional dependence on $\theta_{12}$, thus $\mathbb{E}[\mathcal{F}] = \mathcal{F}$.
% where $\frac{\mathbf{p}_1 \cdot \mathbf{p}_2}{p_1 p_2} = \cos{\theta_{12}}$.
If we assume the best-case scenario, $\mathbb{E}[\cos{\theta_{12}}] = 0$, and the likely scenario, $\mathbb{E}[\cos^2 \theta_{12}] > 0$, then the expected squared error is
\begin{equation}
\begin{aligned}
    \min_{\mathcal{F}} \mathbb{E}[(\mathcal{F} - p^2)^2] &= \min_{\mathcal{F}} \mathcal{F}^2 - 2 \mathcal{F} (p_1^2 + p_2^2) + (p_1^2 + p_2^2)^2 + 4 p_1^2 p_2^2 \mathbb{E}[\cos^2 \theta_{12}] \\
    &= 4 p_1^2 p_2^2 \mathbb{E}[\cos^2 \theta_{12}] > 0;
\end{aligned}
\end{equation}
implying $\mathcal{F}$ is a biased estimator. By introducing rotating features through rotationally equivariant convolutions, the network is effectively introducing functional dependence angles between vector quantities, just like in this example. Given that SchNetPack does so well on the dipole moment, it remains an open question whether distances alone, in an average molecule in QM9, are enough to orient atom-wise contributions to the dipole moment. This investigation is left for future work.

% We note that the situation above is merely for some physical intuition about why including rotating features might be important. A molecule of two atoms could only have an axis-aligned dipole moment. Also, networks which only consider non-rotating features could conceivably reconstruct an approximation angular information of the local environment through repeated application of the triangle inequality. Since these aspects would have to be learned, we hypothesize that it contributes to the parameter efficiency of L1Net.

L1Net is competitive in comparison with the presented architectures despite having fewer parameters and layers than the others. Still, it isn't clear if there is a class of targets which are fundamentally better suited to architectures with rotating features. The ablation study had the most impact on dipole moment $\mu$, electronic spatial extent $\langle R^{2} \rangle$, and energy at 0K $U_{0}$; however, SchNetPack, without rotating features, had the lowest error on those targets and currently holds the state-of-the-art prediction on dipole moment. The SchNet family of models includes 6 convolutions and 20 atom-wise layers (32 including filter generating networks), more than the L$\cdot$Nets. Since our results imply that adding convolutions was not very efficient, it may be that more atom-wise layers are critical to gain expressivity with non-rotating features. We performed a small experiment in this direction within our framework using a network called L0Net Outdeep. See the supplementary material.

Cormorant and L1Net outperformed all architectures without rotating features on isotropic polarizability $\alpha$ and heat capacity $C_{v}$. L1Net includes the gated activation function, while Cormorant does not. Given that L1Net outperformed Cormorant on seven targets, while using fewer parameters, this is evidence that gated nonlinearities are worthwhile. Cormorant used an architecture which could be cast as an L3Net in our framework, by including spherical harmonic features up to degree 3. They applied 4 convolutional layers, ``CGLayers,'' and do not have a clear equivalent to the atom-wise layer. Cormorant includes a so-called ``two-body interaction'' which no other network applies.

\section{Conclusion}

We performed an ablation study of the L1Net in order to determine the value of rotationally equivariant internal layers in regression on molecular properties using the data set QM9. Since other networks like SchNet and Comorant can be cast within our \texttt{e3nn} framework, this experiment provides intuition about architecture design for a wide variety of paradigms. Internal rotationally equivariant layers quantitatively improved performance by 23\% on average while introducing new layers only helped by a mean of 4\%. We provided physical intuition about what is gained by using rotating features using a simple example of a dipole built from two constituents. This example was chosen because the dipole moment was most impacted by the introduction of rotating features compared with increasing depth. However, it remains challenging to identify specific targets which would benefit the most from rotational features, in general.
Our recommendation is to use rotationally equivariant internal layers when performing regression on (magnitudes of) geometric tensors where the angular contribution to constituent addition plays an important role. Our results imply that these contributions play a role in other properties as well, but of lower order.
% ``$\ell=1$ contribution'' is significant.

\paragraph{Acknowledgements}
We acknowledge financial support from the European Commission (ERC CoG 772230 "ScaleCell"), the Berlin Mathematics center MATH+ (AA1-6, EF1-2) and the federal ministry of education and research BMBF (BIFOLD). Tess E. Smidt and Mario Geiger were supported by the Laboratory Directed Research and Development Program of Lawrence Berkeley National Laboratory and Benjamin Kurt Miller was supported by CAMERA both under U.S. Department of Energy Contract No. DE-AC02-05CH11231.
We are grateful for in-depth discussions with Moritz Hoffmann, Kostiantyn Lapchevskyi, Josh Rackers, Jonas Köhler, Jan Hermann, and Simon Batzner. Special thanks to Kostiantyn Lapchevskyi for having a diligent eye.
% TODO add for arxiv
% Kostiantyn Lapchevskyi's diligent eye spotted a critical error in the units of a previous version of our paper which led to significant changes in the text. See the errata section for details.

\paragraph{Code} Our exact implementation can be found at \href{https://github.com/bkmi/equivariant-benchmark}{https://github.com/bkmi/equivariant-benchmark}.

\clearpage

\bibliographystyle{unsrt}
\bibliography{bibliography}

\clearpage

\section{Supplementary Material}

\subsection{Normalization Coefficient}

A feature of \texttt{e3nn} is the convolution defined in Equation \ref{eqn:kc_forward}. One heretofore undefined coefficient in the convolution is $n$. Recall that the normalization coefficient is selected such that component-wise second moment unity input features and component-wise normally distributed radial components produce output features which are component-wise unity in their second moments. 

To better discuss the normalization properties, it makes sense to divide Equation \ref{eqn:kc_forward} into a so-called ``Kernel,''

\begin{equation}
	\label{eqn:kern_forward}
	K^{l_{out}l_{in}}_{ui\;vj} = \sum_{l_f\; k} C^{l_{out}l_{in}l_f}_{ijk} \; Y^{l_f}_k \; R^{l_{out}l_{in}l_f}_{uv} \; n^{l_{out}l_{in}},
\end{equation}

and a ``Kernel-Feature Convolution,''

\begin{equation}
	\label{eqn:conv_forward}
	\Tilde{F}^{l_{out}}_{ui\;A} 
	= \sum_{B\; l_{in}\; vj} K^{l_{out}l_{in}}_{ui\;vj\;AB} \; F^{l_{in}}_{vj\;B}.
\end{equation}

As you can see, calculating the Kernel followed by the Kernel-Feature Convolution yields $\Tilde{F}$ which are the intermediate features before the application of the Gated Block, i.e. $\text{Convolution}(\cdot) = \text{Kernel-Feature Convolution} \circ \text{Kernel}(\cdot)$.

Now that the Kernel is defined, we can discuss the normalization in simpler language. Using the $\langle H \rangle$ notation for the mean of $H$, we write four useful, true statements:

\begin{equation}
\label{eqn:norm1}
\text{(by independence)} \; \; \text{var} \left[ \sum_{l_{in} \, vj} K^{l_{out} l_{in}}_{ui \, vj} F^{l_{in}}_{vj} \right] 
= \sum_{l_{in} \, vj} \text{var} \left[ K^{l_{out} l_{in}}_{ui \, vj} F^{l_{in}}_{vj} \right],
\end{equation}

\begin{equation}
\label{eqn:norm2}
\text{(by independence)} \; \; 
\text{var} \left[ K^{l_{out} l_{in}}_{ui \, vj} F^{l_{in}}_{vj} \right]
= \langle (KF)^2 \rangle - \langle KF \rangle^2
= \langle K^2 \rangle \langle F^2 \rangle - \langle K \rangle^2 \langle F \rangle^2,
\end{equation}

\begin{equation}
\label{eqn:norm3}
(\text{since } \langle R \rangle = 0 ) \; \; \langle K^{l_{out} l_{in}}_{ui \, vj} \rangle
= \sum_{l_f \, k} \langle C^{l_{out}l_{in}l_f}_{ijk} \, Y^{l_f}_k \, R^{l_{out}l_{in}l_f}_{uv} \, n^{l_{out}l_{in}} \rangle 
= 0,
\end{equation}

\begin{equation}
\label{eqn:norm4}
\begin{aligned}
\langle (K^{l_{out} l_{in}}_{ui \, vj})^2 \rangle 
&= \sum_{l_f \, k} \sum_{l_f' \, k'} \langle C^{l_{out}l_{in}l_f}_{ijk} \, C^{l_{out}l_{in}l_f'}_{ijk'} \, Y^{l_f}_k \, Y^{l_f'}_{k'} \, R^{l_{out}l_{in}l_f}_{uv} \, R^{l_{out}l_{in}l_f'}_{uv} \, (n^{l_{out}l_{in}})^2 \rangle \\
(\text{since } \langle R R \rangle = \delta) 
&= \sum_{l_{f}} \sum_{k k'} C^{l_{out}l_{in}l_f}_{ijk} \, C^{l_{out}l_{in}l_f}_{ijk'} \, Y^{l_f}_k \, Y^{l_f}_{k'} \, (n^{l_{out}l_{in}})^2 \\
&= (n^{l_{out}l_{in}})^2 \sum_{l_f} \left( \sum_k  C^{l_{out}l_{in}l_f}_{ijk}\, Y^{l_f}_k \right)^2.
\end{aligned}
\end{equation}

Now we can combine Equations \ref{eqn:norm3} and \ref{eqn:norm4} with Equation \ref{eqn:norm2} to write,

\begin{equation}
\text{var} \left[ K^{l_{out} l_{in}}_{ui \, vj} F^{l_{in}}_{vj} \right] 
= (n^{l_{out}l_{in}})^2 \langle (F^{l_{in}}_{vj})^2 \rangle \sum_{l_f} \left( \sum_k  C^{l_{out}l_{in}l_f}_{ijk}\, Y^{l_f}_k \right)^2.
\end{equation}

This implies that Equation \ref{eqn:norm1} can be written

\begin{equation}
\begin{aligned}
\text{(\ref{eqn:norm1})} 
&= \sum_{l_{in}} (n^{l_{out}l_{in}})^2 \sum_{vj} \tau^2_{l_{in}} \sum_{l_f} \left( \sum_k  C^{l_{out}l_{in}l_f}_{ijk}\, Y^{l_f}_k \right)^2 \\
&= \sum_{l_{in}} \left( \sum_{v} 1 \right) (n^{l_{out}l_{in}})^2 \tau^2_{l_{in}} \sum_{l_f} \sum_{j} \left( \sum_k  C^{l_{out}l_{in}l_f}_{ijk}\, Y^{l_f}_k \right)^2 \\
&= \sum_{l_{in}} \left( \sum_{v} 1 \right) (n^{l_{out}l_{in}})^2 \tau^2_{l_{in}} \sum_{l_f} (4 \pi (2 l_{out} + 1))^{-1} \\
&= (4 \pi (2 l_{out} + 1))^{-1} \sum_{l_{in}} \left( \sum_{v} 1 \right) (n^{l_{out}l_{in}})^2 \, \tau^2_{l_{in}} \left( \sum_{l_f} 1 \right),
\end{aligned}
\end{equation}

where $\langle (F^{l_{in}}_{vj})^2 \rangle \coloneqq \tau^2_{l_{in}}$. Note that we want (\ref{eqn:norm1}) $= \tau^2_{l_{in}}$, and we assume that $\tau^2_{l_{in}} = 1$. This enforces that the second moment is unity. Therefore,

\begin{equation}
\begin{aligned}
4 \pi (2 l_{out} + 1) 
&= \sum_{l_{in}} \left( \sum_{v} 1 \right) (n^{l_{out}l_{in}})^2 \left( \sum_{l_f} 1 \right) \\
(n^{l_{out}l_{in}})^2
&= \frac{4 \pi (2 l_{out} + 1)}{\left( \sum_{v} 1 \right) \left( \sum_{l_f} 1 \right) \left( \sum_{l_{in}} 1 \right) }.
\end{aligned}
\end{equation}

\subsection{Shift \& Scale Function}
\label{sec:shift_and_scale}

Neural networks operate best when their outputs are normally distributed. For this reason, we perform a decomposition of the target value such that the regression network's output fits this criteria. The implementation of this part of the network was handled by the SchNetPack package \cite{schnetpack}. First, the decomposition utilizes the reference values in the QM9 data set so the network starts with a good guess and predicts a perturbation from that guess. The network's prediction is decomposed into a reference bias, an atom-wise sum from the Table \ref{table:atomref}, and a scaled contribution from each atom. 

\begin{table}[hbt]
	\centering
	\resizebox{\textwidth}{!}{
		\begin{tabular}{c||c|c|c|c|c|c}
			Element & ZPVE &  U (0 K) & U (298.15 K) & H (298.15 K) & G (298.15 K) & Heat Capacity \\
			& Hartree & Hartree & Hartree & Hartree & Hartree & Cal/(Mol Kelvin) \\
			\hline\hline
			H & 0.000 &  -0.500 &  -0.499 &  -0.498 &  -0.511 & 2.981 \\
			C & 0.000 & -37.847 & -37.845 & -37.844 & -37.861 & 2.981 \\
			N & 0.000 & -54.584 & -54.582 & -54.582 & -54.599 & 2.981 \\
			O & 0.000 & -75.065 & -75.063 & -75.062 & -75.080 & 2.981 \\
			F & 0.000 & -99.719 & -99.717 & -99.716 & -99.734 & 2.981
		\end{tabular}
	}
	\caption{\label{table:atomref} Table is adapted from the ``atom ref'' table in the QM9 paper \cite{qm9}.}
\end{table}

For any target in Table \ref{table:atomref} and atom in QM9, we can create a map from element $Z$ and prediction column $C$ to the corresponding reference value $ref(Z, C)$. For example, $ref(\text{H}, \text{U}_0) = -0.5$ Hartree. Given a training set of $M$ molecules indexed by $m \in \{1, 2, ..., M\}$ each with $A_m$ atoms indexed by $a_m \in \{1, 2, ..., A_m\}$ with a corresponding element $Z_{a_m}$, we write the reference bias

\begin{equation}
p_m = \sum_{a_m=1}^{A_m} ref(Z_{a_m}, C).
\end{equation}

To further our decomposition consider the target regression value for a certain molecule $t_m$. From the ground truth, we can write the atom-wise deviation from the reference value,

\begin{equation}
\tilde{t}_m = \frac{t_m - p_m}{A_m}.
\end{equation}

By gathering statistics from the training data on this $\tilde{t}_m$, we will achieve our goal of normalizing the output of the regression network. Let  $\bar{\tilde{t}}$ and $\sigma_{\tilde{t}}$ be the mean and standard deviation of $\tilde{t}_m$ over molecules respectively. Given the atom-wise output of a regression network $\mathcal{R}_{a_m}$, we predict the ground truth target $\hat{t}_m$ by

\begin{equation}
\begin{aligned}
\label{eqn:schnetpack_output_predict}
\hat{t}_m &= p_m + \sum_{a_m=1}^{A_m} \left( \bar{\tilde{t}} + \sigma_{\tilde{t}} \, \mathcal{R}_{a_m} \right) \\
&= p_m + A_m \bar{\tilde{t}} + \sigma_{\tilde{t}} \sum_{a_m}^{A_m} \mathcal{R}_{a_m} \\
&= p_m + \frac{A_m}{M} \sum_{n=1}^{M} \tilde{t}_n + \sigma_{\tilde{t}} \sum_{a_m}^{A_m} \mathcal{R}_{a_m}. \\
\end{aligned}
\end{equation}

We presented several equivalent formulations in order to provide clarity.

\subsection{Hyperparameter Search Technique}

The technique applied in the paper was to do an ablation study of the rotating features in an L1Net but first we had to determine which hyperparameters defined the L1Net. In order to find a network architecture which was well suited for every QM9 target, the hyperparameter search utilized multi-target training using a featurization-output design. By searching in the multi-target regime, as opposed to doing 12 individual searches utilizing the same architecture, we traded the accuracy of single-target training for a factor of 12 decrease in hyperparameter search time. This allowed for significantly more architectures to be tested.

\begin{figure}[htb]
	\centering
	\includegraphics[width=0.54\textwidth]{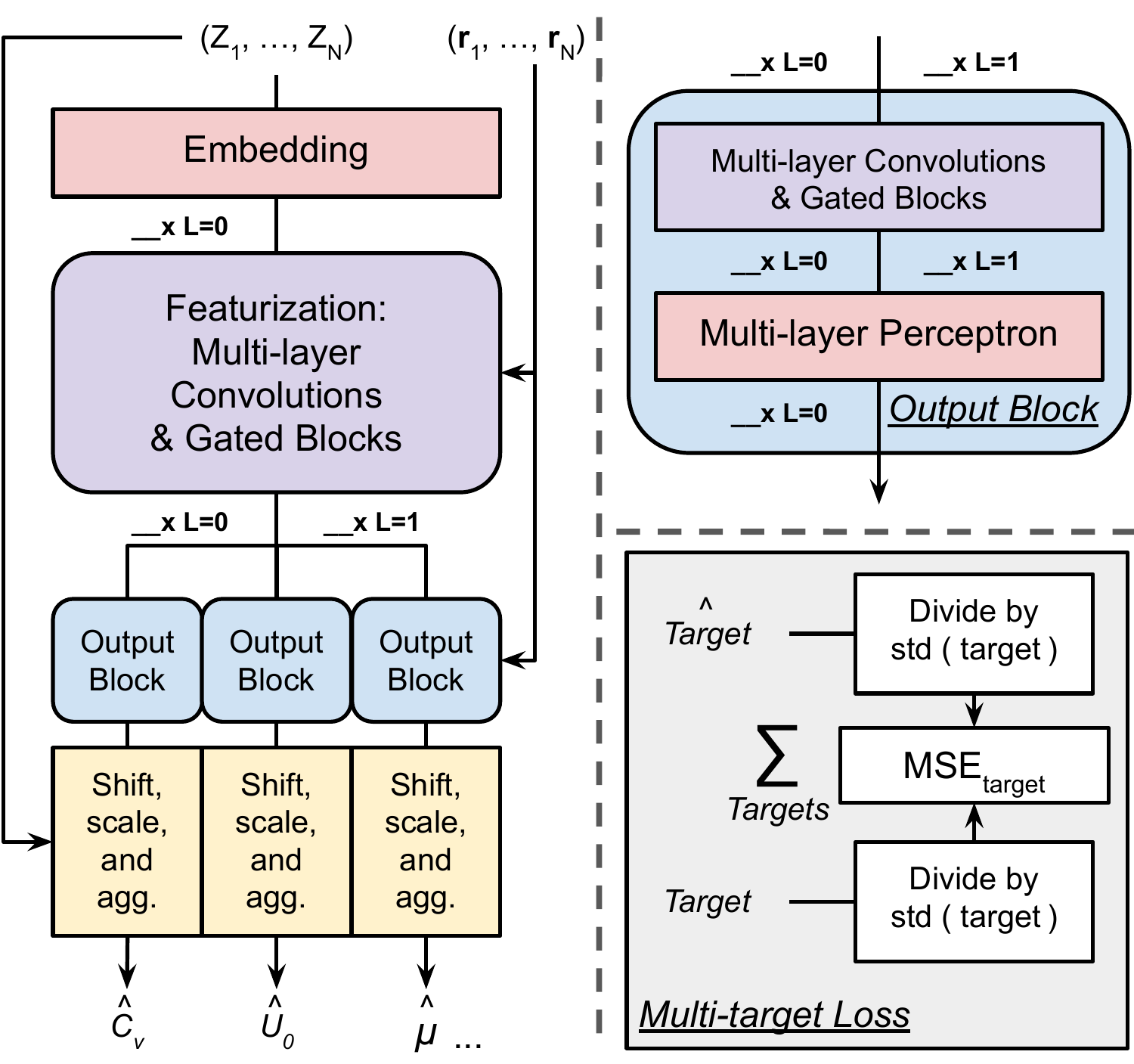}
	\caption{Illustration of the hyperparameter search with the set of possible architectures on the left, the set of possible output blocks on the top right, and the multi-target, normalized loss function shown in the bottom right.}
	\label{fig:e3nn_hp_search}
\end{figure}

The featurization section in L1Net (see Figure \ref{fig:e3nn_l0_architecture}) is represented by the atom-wise embedding and two convolution \& gated block layers. In L1Net the so-called output block represents the remaining convolution \& gated block layer. Each final atom-wise layer and shift, scale, and aggregate is unique to the target being predicted. Given that examples, we focus next on how we achieved the multi-target training aspect using multiple output blocks.

Each output block receives a copy of the same learned featurization; however, the output blocks do not share gradients or other information. This allows for each output block to transform the learned input features in parallel, each predicting a single target; thereby, the whole network makes predictions on multiple targets. The network design is depicted in Figure \ref{fig:e3nn_hp_search}.

The featurization section is an embedding followed by layers of Convolutions and Gated Blocks. A combination of order zero and order one spherical harmonic features are copied and passed to each output block. Each output block then passes those features through Convolutions and Gated Blocks and calculates an array of scalar features. Since they are scalar features, we can pass them through one or more atom-wise layers with a rectified linear unit activation without breaking total network rotation invariance. The last layer of that multi-layer perceptron predicts a single scalar using an atom-wise layer with the identity activation function, which is then passed to the shift and scale operation as seen in Section \ref{sec:shift_and_scale}.

The loss calculation is different from other learning problems because we attempt to normalize the losses across targets. Since all targets are equally important, we normalize their variance to one based off of statistics from our training data. This formulation depends on the assumption that each output block predicts mean zero at initialization. Recall Equation \ref{eqn:schnetpack_output_predict}.

Therefore, using the notation from Section \ref{sec:shift_and_scale}, we write the loss using the the total molecule-wise offset $s_m = p_m + A_m \bar{\tilde{t}}$. The molecule-wise loss for target $t_m$ looks like,

\begin{equation}
\mathcal{L}_m (t_m, \hat{t}_m) = (\frac{t_m - s_m}{\sigma_{\tilde{t}}} - \frac{\hat{t}_m - s_m}{\sigma_{\tilde{t}}})^2 = \frac{1}{\sigma_{\tilde{t}}^2}(t_m - \hat{t}_m - 2 s_m)^2.
\end{equation}

For a batch of molecules $M$ and pairs of targets with corresponding predictions $\{(t_m, \hat{t}_m) : m \in M\}$, the total loss is calculated by

\begin{equation}
\frac{1}{M} \sum_{(t_m, \, \hat{t}_m)} \sum_{m} \mathcal{L}_m (t_m, \hat{t}_m).
\end{equation}

Although it is possible to train a model against all targets at the same time using this methodology, it is often much more difficult to achieve simultaneously good performance across targets. Therefore this model was only used for hyperparameter search, not for reported performance results.

%We choose hyperparameters randomly from within the ranges shown in Table \ref{table:hyperparams_rand_ranges}. 
%
%The choice of radial basis is determined by selecting one of the Cosine, Gaussian, and Bessel bases with one third probability for each. The different bases are defined in B.K. Miller's thesis \cite{ben_thesis}; however, the Gaussian basis came from SchNet \cite{schnet} and the Bessel basis came from DimeNet \cite{dimenet}. The representation, i.e. amount of order zero spherical harmonics and order one spherical harmonics per convolutional layer, is determined like so: Let $0 \leq p \leq 1$ be a random number and $C$ is the number of components, the multiplicity of order zero $R_0$ and order one $R_1$ spherical harmonics are determined by
%
%\begin{equation}
%\label{eqn:hp_distribute_components}
%R_0 = p C, \; \; \; \; R_1 = (1-p)C.
%\end{equation}

\begin{table}[htb]
	\centering
	\begin{tabular}{c||c|c}
		Hyperparameter & Minimum & Maximum \\
		\hline \hline
		Batch Size & 8 & 25 \\
		Learning Rate & $10^{-6}$ & $3 \times 10^{-1}$ \\
		Size of Embedding & 80 & 144 \\
		Featurization Components (FC) & 80 & 144 \\
		Featurization Representation & (FC) randomly divided & between $Y^0_m$ and $Y^1_m$ \\
		Featurization Conv \& GBs & 2 & 5 \\
		Residual Network & True & True \\
		Radial Basis & $\phi_C, \phi_G, \phi_B$ & $\phi_C, \phi_G, \phi_B$ \\
		Number of Radial Bases & 25 & 100 \\
		Radial Maximum & 1.2 \AA & 30.0 \AA \\
		Radial MLP Layers & 1 & 3 \\
		Radial MLP Neurons & 80 & 144 \\
		Output Components (OC) & 64 & 128 \\
		Output Representation & (OC) randomly divided & between $Y^0_m$ and $Y^1_m$ \\
		Output Conv \& GBs & 1 & 2 \\
		Output MLP Layers & 1 & 3 \\
		Output MLP Neurons & 80 & 144
	\end{tabular}
	\caption{The ranges of hyperparameters for the random hyperparameter search are written in this table. Cosine $\phi_C$, Gaussian $\phi_G$, and Bessel $\phi_B$ are defined in Equation \ref{eqn:cosine_basis}, SchNet \cite{schnet}, or DimeNet \cite{dimenet} respectively.}
	\label{table:hyperparams_rand_ranges}
\end{table}

The hyperparameter search involved sampling forty different sets of hyperparameters from the ranges in Table \ref{table:hyperparams_rand_ranges} and doing multi-target training for ten epochs with each set of hyperparameters. The test set performance for each of the forty models was compared. We took L1Net to be the winner since it produced the minimum loss averaged over normalized losses on all targets.

\subsection{L1Net, L0Net, etc. Learning Plots}

\begin{figure}[htb]
	\centering
	\includegraphics[width=0.9\textwidth]{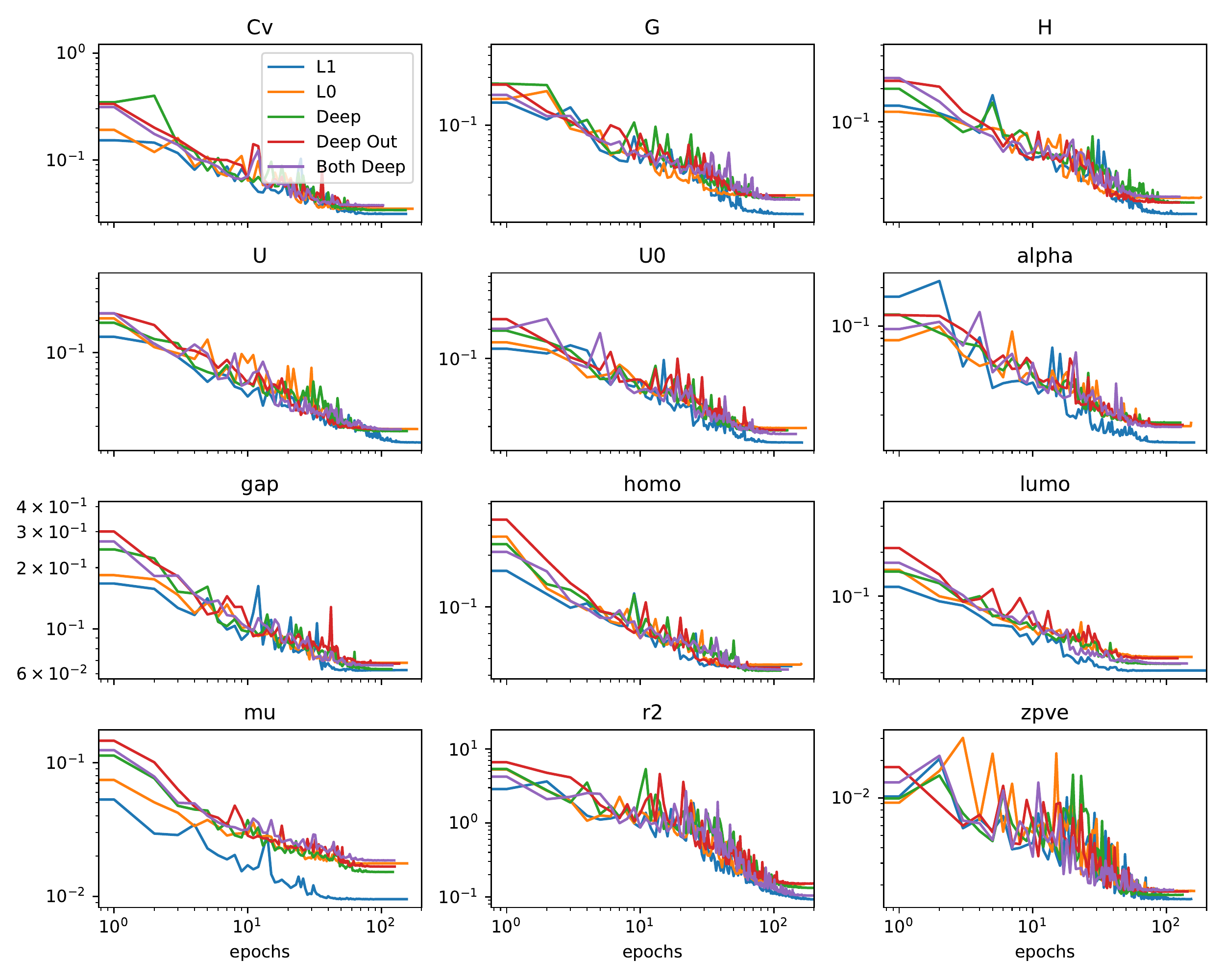}
	\caption{Plotted above is the logarithm of the mean absolute error on the validation set versus the logarithm of training epochs for every regression target. The plots contain the training curves for the L1Net, L0Net, L0Net Deep, L0Net Outdeep, and L0Net Both Deep architectures. Just like in the main article, the adam optimizer was employed with standard parameters and an initial learning rate of $6.53 \times 10^{-3}$. The learning rate was decayed given a loss plateau of five epochs to a minimum of $10^{-7}$. The maximum number of training epochs was set at 200 with early stopping patience of 50.}
	\label{fig:mae_vs_training}
\end{figure}

We compared the performance of L1Net to several different L0Net-style architectures. The most important question in this paper was: ``Can an L0Net make-up for the L1Net performance by increasing depth?'' However, given our architecture design, ``increasing depth'' could mean one of several things. L0Net Deep increased added another Convolution \& Gated Block layer, L0Net Outdeep added another atom-wise layer after the convolutions, and L0Net Both Deep did both of those things. Their performance on validation data is plotted in Figure \ref{fig:mae_vs_training}. We found that the L0Net Deep performed the best when compared with the other L0Net-style architectures.

% TODO for arxiv
% \subsection{Errata} A previous version of this work included incorrect quantitative data regarding the performance of L1Net, L0Net, and L0Net Deep on $\mu$, $\alpha$, and $\langle R^2 \rangle$. Despite reporting the units of these values as Debye, $a_{0}^{3}$, and $a_{0}^{2}$ respectively, our values were actually written in $e$\AA, \AA$^{3}$, and \AA$^{2}$ respectively, where $e$ implies the charge of an electron. This quantitative error led to incorrect assumptions about the performance of our L1Net. The error was found by Kostiantyn Lapchevskyi in his efforts to replicate our results. The previous version of our paper can be found on arxiv in version 1 and 2.

\end{document}